\documentclass[10pt, a4paper]{article}
\usepackage{lrec}
\usepackage{graphicx}
\usepackage[table]{xcolor}
\usepackage{tabularx}
\usepackage{soul}
\usepackage[small,bf]{caption}
\usepackage{booktabs}
\usepackage{epstopdf}
\usepackage{url}
\usepackage{multirow}
\usepackage{multicol}

\usepackage[hidelinks]{hyperref}
\usepackage{xstring}

\makeatletter
\renewcommand{\p@section}{\arabic{section}\expandafter\@gobble}
\renewcommand{\p@subsection}{\thesection\arabic{subsection}\expandafter\@gobble}
\renewcommand{\p@subsubsection}{\thesubsection\arabic{subsubsection}\expandafter\@gobble}
\makeatother

\title{Common Voice: A Massively-Multilingual Speech Corpus}

\name{Rosana Ardila,$^{\dagger}$ Megan Branson,$^{\dagger}$ Kelly Davis,$^{\dagger}$ Michael Henretty, Michael Kohler, Josh Meyer,$^{\circ}$\\[1mm]
\textbf{{\large Reuben Morais,$^{\dagger}$ Lindsay Saunders,$^{\dagger}$ Francis M. Tyers,$^{\ddagger}$ Gregor Weber$^{\dagger}$}}
}

\address{$^{\dagger}$ Mozilla~~~$^{\ddagger}$ Indiana University~~~$^{\circ}$ Artie, Inc.\\
  Various Cities ~~~ Bloomington, IN, USA ~~~ Los Angeles, CA, USA \\
  \{rosana, mbranson, kdavis, reuben, lsaunders, gweber\}@mozilla.com, \\
  ftyers@iu.edu, michael.henretty@gmail.com, me@michaelkohler.info, josh.meyer@artie.com \\
}

\abstract{
  The Common Voice corpus is a massively-multilingual collection of transcribed speech intended for speech technology research and development. Common Voice is designed for Automatic Speech Recognition purposes but can be useful in other domains (e.g. language identification).
  To achieve scale and sustainability, the Common Voice project employs crowdsourcing for both data collection and data validation. The most recent release includes 29 languages, and as of November 2019 there are a total of 38 languages collecting data. Over 50,000 individuals have participated so far, resulting in 2,500 hours of collected audio.
  To our knowledge this is the largest audio corpus in the public domain for speech recognition, both in terms of number of hours and number of languages.
  As an example use case for Common Voice, we present speech recognition experiments using Mozilla's DeepSpeech Speech-to-Text toolkit\footnote{\url{https://github.com/mozilla/DeepSpeech/tree/v0.3.0}}. By applying transfer learning from a source English model, we find an average Character Error Rate improvement of $5.99 \pm 5.48$ for twelve target languages (German, French, Italian, Turkish, Catalan, Slovenian, Welsh, Irish, Breton, Tatar, Chuvash, and Kabyle). 
  For most of these languages, these are the first ever published results on end-to-end Automatic Speech Recognition.
 \newline  }

\begin{document}

\maketitleabstract

\section{Introduction}

The Common Voice project\footnote{\url{http://voice.mozilla.org}} is a response to the current state of affairs in speech technology, in which training data is either prohibitively expensive or unavailable for most languages \cite{roter_2019}. We believe that speech technology (like all technology) should be open and decentralized, and the Common Voice project achieves this goal via a mix of community building, open source tooling, and a permissive licensing scheme. The corpus is designed to organically scale to new languages as community members use the provided tools to translate the interface, submit text sentences, and finally record and validate voices in their new language \footnote{\url{https://discourse.mozilla.org/t/readme-how-to-see-my-language-on-common-voice/31530}}. The project was started with an initial focus on English in July 2017 and then in June 2018 was made available for any language.

The remainder of the paper is organized as follows: In Section (\ref{sec:motivation}) we motivate Common Voice and review previous multilingual corpora. Next, in Section (\ref{sec:process}) we describe the recording and validation process used to create the corpus. Next, in Section (\ref{sec:contents}) we describe the current contents of Common Voice, and lastly in Section (\ref{sec:experiments}) we show multilingual Automatic Speech Recognition experiments using the corpus.

\section{Prior work}\label{sec:motivation}
Some notable multilingual speech corpora include VoxForge \cite{voxforge}, Babel \cite{gales2014speech}, and M-AILABS \cite{mailabs}. Even though the Babel corpus contains high-quality data from 22 minority languages, it is not released under an open license. VoxForge is most similar to Common Voice in that it is community-driven, multilingual (17 languages), and released under an open license (GNU General Public License). However, the VoxForge does not have a sustainable data collection pipeline compared to Common Voice, and there is no data validation step in place. M-AILABS data contains 9 language varieties with a modified BSD 3-Clause License, however there is no community-driven aspect. Common Voice is a sustainable, open alternative to these projects which allows for collection of minority and majority languages alike.

\section{Corpus Creation}\label{sec:process}

\begin{figure}[!htbp]
	\centering
\fbox{
	\includegraphics[width=.4\textwidth]{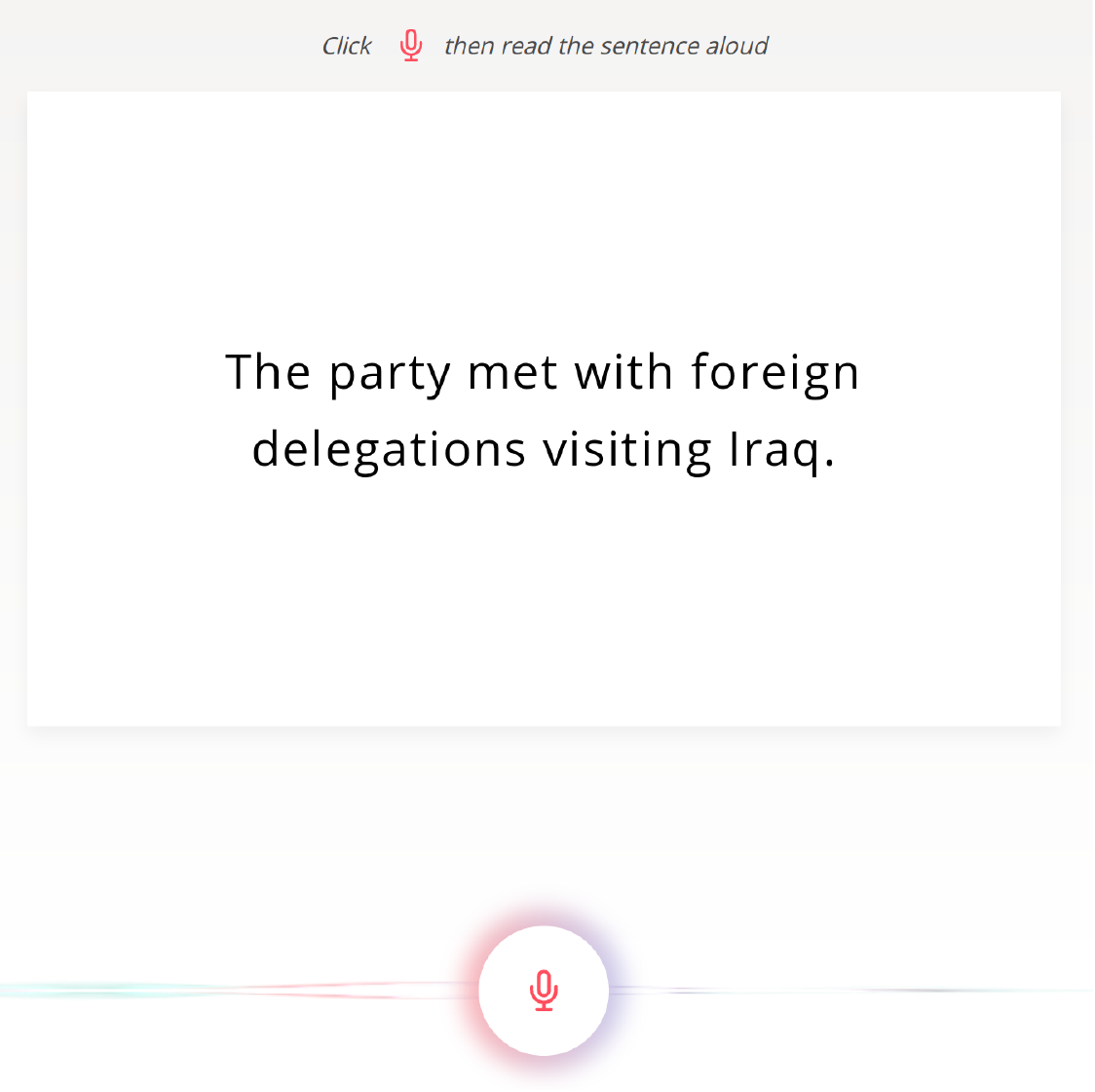}
}
	\caption{Recording interface for Common Voice. Additionally, it is possible to skip or report as problematic any audio or sentence.}
	\label{fig:speak}
\end{figure}

\begin{figure}[!htbp]
	\centering
\fbox{
	\includegraphics[width=.4\textwidth]{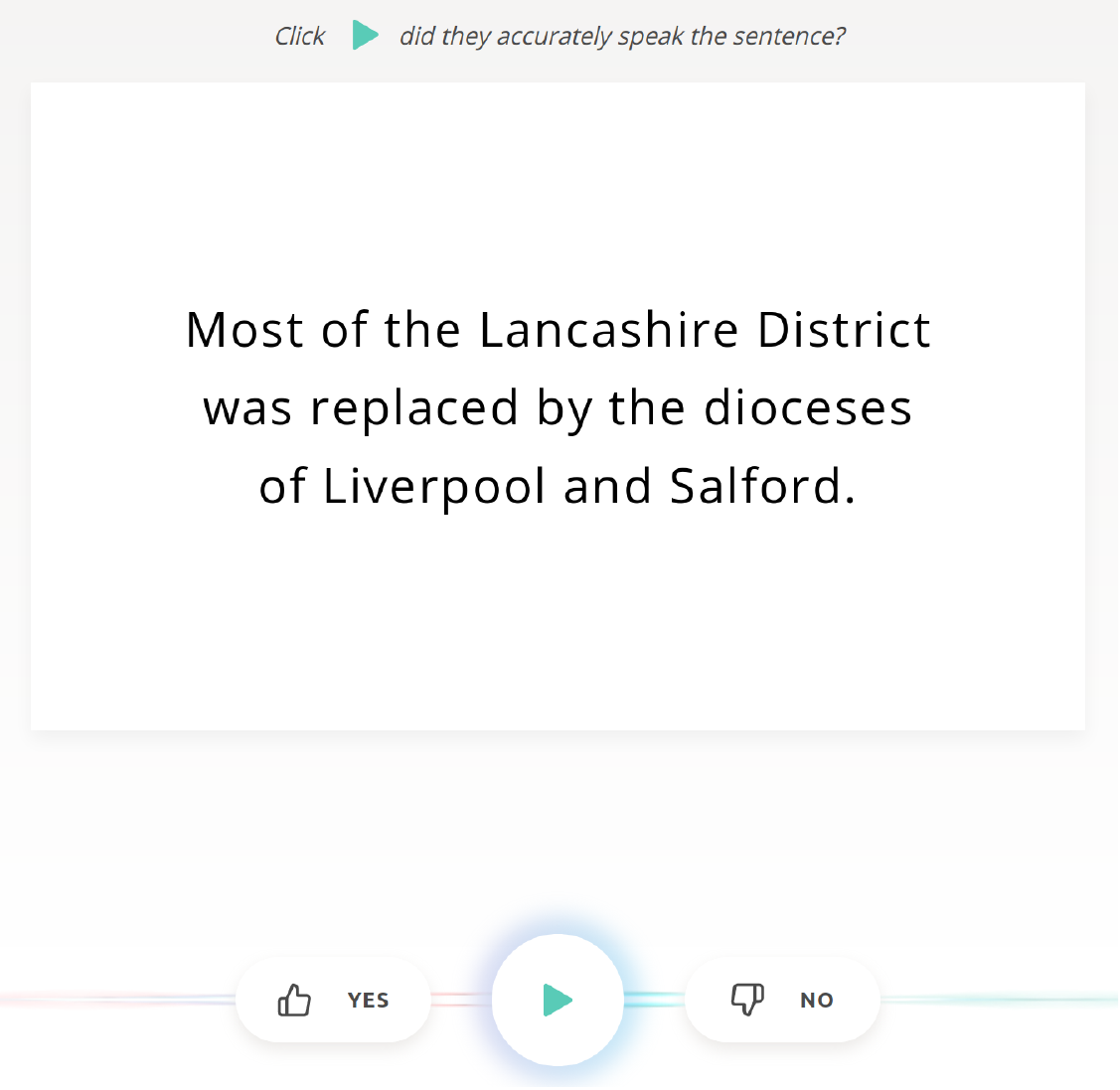}
}
	\caption{Validation interface for Common Voice. Additionally, it is possible to skip or report as problematic any audio or sentence.}
	\label{fig:listen}
\end{figure}

The data presented in this paper was collected and validated via Mozilla's Common Voice initiative. Using either the Common Voice website or iPhone app, contributors record their voice by reading sentences displayed on the screen (see Figure (\ref{fig:speak})). The recordings are later verified by other contributors using a simple voting system. Shown in Figure (\ref{fig:listen}), this validation interface has contributors mark $<$audio,transcript$>$ pairs as being either correct (up-vote) or incorrect (down-vote).

A maximum of three contributors will listen to any audio clip.\footnote{In the early days of Common Voice, this voting mechanism contained bugs, and some clips in the official release received over three votes. In these cases we use a simple majority rule. The total number of up and down votes is released with the dataset.} If an $<$audio,transcript$>$ pair first receives two up-votes, then the clip is marked as valid. If instead the clip first receives two down-votes, then it is marked as invalid. A contributor may switch between recording and validation as they wish.

\begin{figure}[!htbp]
	\centering
\fbox{
	\includegraphics[width=.4\textwidth]{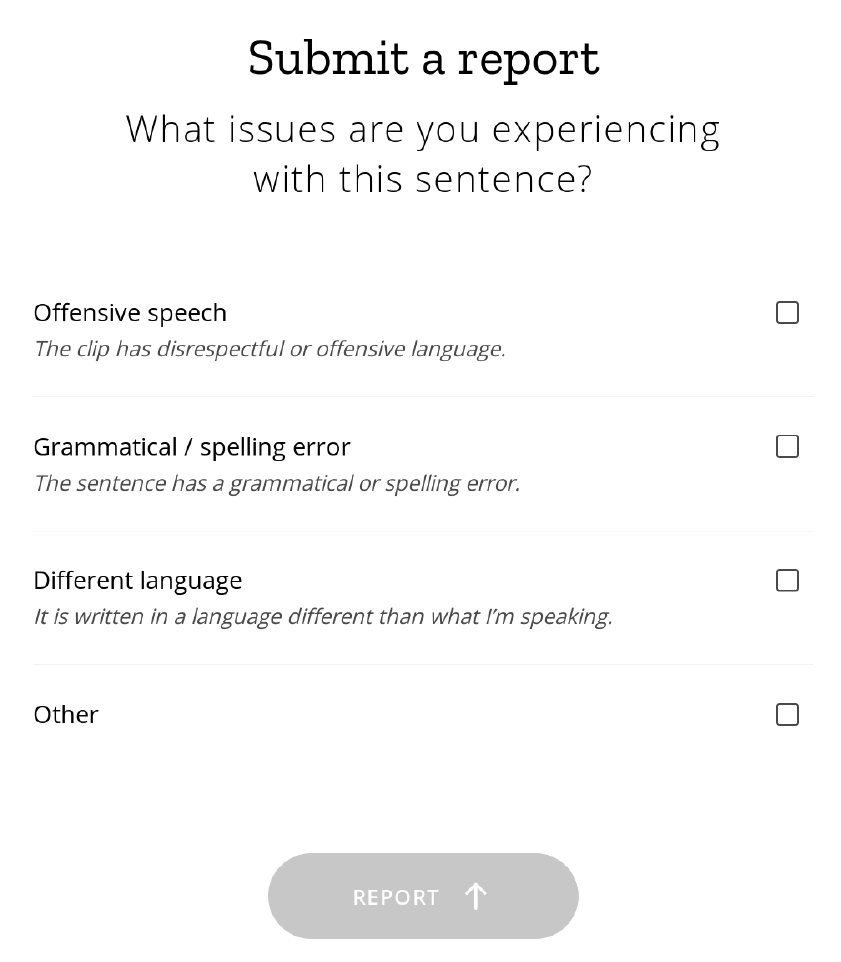}
}
	\caption{Interface for reporting problematic data. It is possible to report any text or audio as problematic during either recording or validation.}
	\label{fig:problematic}
\end{figure}

Only clips marked as valid are included in the official training, development, and testing sets for each language. Clips which did not receive enough votes to be validated or invalidated by the time of release are released as ``other''. The train, test, and development sets are bucketed such that any given speaker may appear in only one. This ensures that contributors seen at train time are not seen at test time, which would skew results. Additionally, repetitions of text sentences are removed from the train, test, and development sets of the corpus.\footnote{Repetitions may be found in the other TSV files included in Common Voice.}

The number of clips is divided among the three datasets according to statistical power analyses. Given the total number of validated clips in a language, the number of clips in the test set is equal to the number needed to achieve a confidence level of 99\% with a margin of error of 1\% relative to the number of clips in the training set. The same is true of the development set.\footnote{All code which performs the bucketing can be found here: \url{https://github.com/mozilla/CorporaCreator}}

The audio clips are released as mono-channel, 16bit MPEG-3 files with a 48kHz sampling rate. The choice to collect and release MPEG-3 as opposed to a lossless audio format (e.g. WAV) is largely due to the web-based nature of the Common Voice collection platform. MPEG-3 is the most universally supported audio format for the web, and as such is the most reliable recording/playback technique for various devices and browsers. Also practically speaking, the audio quality is appropriate for speech applications.

\section{Corpus Contents}\label{sec:contents}
\subsection{Released Languages}

The data presented in Table (\ref{tab:cv-june}) shows the currently available data. Each of the released languages is available for individual download as a compressed directory from the Mozilla Common Voice website.\footnote{Common Voice Download page: \url{https://voice.mozilla.org/en/datasets}} The directory contains six files with Tab-Separated Values (i.e. TSV files), and a single \texttt{clips} sub-directory which contains all of the audio data. Each of the six TSV files represents a different segment of the voice data, with all six having the following column headers: [\texttt{client\_id, path, sentence, up\_votes, down\_votes, age, gender, accent}]. The first three columns refer to an anonymized ID for the speaker, the location of the audio file, and the text that was read. The next two columns contain information on how listeners judged the $<$audio,transcript$>$ pair. The last three columns represent demographic data which was optionally self-reported by the speaker of the audio.

\begin{table}
  \centering
\scalebox{0.9}{
\begin{tabular}{llrrrr}
\toprule
\multirow{2}{*}{\textbf{Language}}&\multirow{2}{*}{\textbf{Code}}&\multirow{2}{*}{\textbf{Voices}}&\multicolumn{2}{c}{\textbf{Hours}}\\
&&&Total&Validated\\
\midrule
\emph{Abkhaz}&\texttt{ab}&3&$<$1&$<$1\\
\emph{Arabic}&\texttt{ar}&225&15&9\\
Basque&\texttt{eu}&508&83&46\\
Breton&\texttt{br}&118&10&3\\
Catalan&\texttt{ca}&1,834&120&107\\
Chinese (China)&\texttt{zh-ZH}&288&12&11\\
Chinese (Taiwan)&\texttt{zh-TW}&949&43&33\\
Chuvash&\texttt{cv}&38&2&1\\
Dhivehi&\texttt{dv}&92&8&5\\
Dutch&\texttt{nl}&502&23&18\\
English&\texttt{en}&39,577&1,087&780\\
Esperanto&\texttt{eo}&129&16&13\\
Estonian&\texttt{et}&225&12&11\\
French&\texttt{fr}&3,005&184&173\\
German&\texttt{de}&5,007&340&325\\
Hakha Chin&\texttt{cnh}&280&4&2\\
\emph{Indonesian}&\texttt{id}&54&5&4\\
\emph{Interlingua}&\texttt{ia}&11&2&1\\
Irish&\texttt{ga}&63&3&2\\
Italian&\texttt{it}&602&40&36\\
\emph{Japanese}&\texttt{ja}&48&2&1\\
Kabyle&\texttt{kab}&584&192&181\\
Kinyarwanda&\texttt{rw}&32&1&$<$1\\
Kyrgyz&\texttt{ky}&97&20&8\\
\emph{Latvian}&\texttt{lv}&82&8&6\\
Mongolian&\texttt{mn}&230&9&8\\
Persian&\texttt{fa}&1,240&70&67\\
\emph{Portuguese}&\texttt{pr}&316&30&27\\
Russian&\texttt{ru}&64&31&27\\
Sakha&\texttt{sah}&35&6&3\\
Slovenian&\texttt{sl}&42&5&2\\
Spanish&\texttt{es}&611&31&27\\
Swedish&\texttt{sv}&44&3&3\\
\emph{Tamil}&\texttt{ta}&89&5&3\\
Tatar&\texttt{tt}&132&26&22\\
Turkish&\texttt{tr}&344&10&9\\
\emph{Votic}&\texttt{vot}&2&$<$1&$<$1\\
Welsh&\texttt{cy}&748&48&42\\
\midrule
\textsc{Total} & &58,250&	2,508&	2,019\\
\bottomrule
\end{tabular}
}

\caption{Current data statistics for Common Voice. Data in \emph{italics} is as of yet unreleased. Other numbers refer to the data published in the June 12, 2019 release.}\label{tab:cv-june}
\end{table}

\subsection{Adding a new Language}

In order to add a new language to the Common Voice project, two steps must be completed. First, the web-app user interface must be translated into the target language. For example, the text shown in Figure (\ref{fig:problematic}) must be translated. Secondly, text prompts must be gathered in order to be read aloud. These texts are not translated, but gathered from scratch for each language -- translation would be very slow and not scalable. 

Translation of the interface is managed through the Pontoon platform\footnote{Pontoon translation platform: \url{https://pontoon.mozilla.org/projects/common-voice/}}. Pontoon allows community members to propose translations, and then the moderators for that language approve or decline the proposals. At the time of writing this paper there are 610 text strings used in the Common Voice interface, where each string can range in length from an isolated word to a paragraph of text.

Collecting text for reading aloud is the second step of adding a new language to Common Voice. For languages with more than 500,000 Wikipedia articles, text sentences are extracted from Wikipedia using community provided rule-sets per language\footnote{Code for Wikipedia extraction can be found here: \url{https://github.com/Common-Voice/common-voice-wiki-scraper}}. These sentences make up the initial text prompts for the languages in question.

Any language community can gather additional sentences through the Sentence Collector\footnote{Sentence Collector web app: \url{https://common-voice.github.io/sentence-collector}} taking advantage of automatic validation mechanisms such as checks for sentence length, foreign alphabets, and numbers. Every sentence submitted through the Sentence Collector needs to be approved by two out of three reviewers, leading to a weekly export of new sentences into the Common Voice database. Once the website is translated and at least 5,000 sentences have been added, the language is enabled for voice recordings.

\section{Automatic Speech Recognition Experiments}\label{sec:experiments}

The following experiments demonstrate the potential to use the Common Voice corpus for multilingual speech research. These results represent work on an internal version of Common Voice from February 2019. The current corpus contains more languages and more data per language.

These experiments use an End-to-End Transfer Learning approach which bypasses the need for linguistic resources or domain expertise \cite{meyer2019multi}. Certain layers are copied from a pre-trained English source model, new layers are initialized for a target language, the old and new layers are stitched together, and all layers are fine-tuned via gradient descent.

\subsection{Data}

We made dataset splits (c.f. Table (\ref{table:data})) such that one speaker's recordings are only present in one data split. This allows us to make a fair evaluation of speaker generalization, but as a result some training sets have very few speakers, making this an even more challenging scenario. The splits per language were made as close as possible to 80\% train, 10\% development, and 10\% test.\footnote{These experiments were performed before Common Voice was in it's current form. As such, the train, development, and test splits here do not correspond exactly to the official releases.}

\begin{table}
  \centering
\scalebox{0.8}{
\begin{tabular}{llrrrrrr}
 \toprule
 & & \multicolumn{6}{c}{\textbf{Dataset Size}} \\
\multirow{3}{*}{\textbf{Language}} & \multirow{3}{*}{\textbf{Code}} & \multicolumn{3}{c}{Audio Clips} & \multicolumn{3}{c}{Unique Speakers} \\
    &      & Dev & Test & Train & Dev & Test & Train  \\
\midrule
Slovenian& \texttt{sl} & 110 & 213 & 728 &1 & 12 & 3 \\
Irish& \texttt{ga} & 181 & 138 & 1,001 & 4& 12 & 6 \\
Chuvash& \texttt{cv} & 96 & 77 & 1,023 &4 & 12 & 5 \\
Breton& \texttt{br} & 163 & 170 & 1,079 & 3& 15 & 7 \\
Turkish& \texttt{tr} & 407 & 374 & 3,771 &32 & 89 & 32 \\
Italian& \texttt{it} & 627 & 734 & 5,019 &29 & 136 & 37 \\
Welsh& \texttt{cy} & 1,235 & 1,201 & 9,547 & 51 & 153 & 75 \\
Tatar& \texttt{tt} & 1,811 & 1,164 & 11,187 & 9& 64 & 3 \\
Catalan& \texttt{ca} & 5,460 & 5,037 & 38,995 &286 & 777 & 313 \\
French& \texttt{fr} & 5,083 & 4,835 & 40,907 & 237 & 837 & 249 \\
Kabyle& \texttt{kab} & 5,452 & 4,643 & 43,223 & 31 & 169 & 63 \\
German& \texttt{de} & 7,982 & 7,897 & 65,745  & 247 & 1,029  & 318  \\
\bottomrule
\end{tabular}
}
  \caption{Data used in the experiments, from an earlier multilingual version of Common Voice. Number of audio clips and unique speakers.}\label{table:data}
\end{table}

Results from this dataset are interesting because the text and audio are challenging, the range of languages is wider than any openly available speech corpus, and the amount of data per language ranges from very small (less than 1,000 clips for Slovenian) to relatively large (over 65,000 clips for German).

\subsection{Model architecture}

\begin{figure}[!htbp]
  \centering
\includegraphics[width=.4\textwidth]{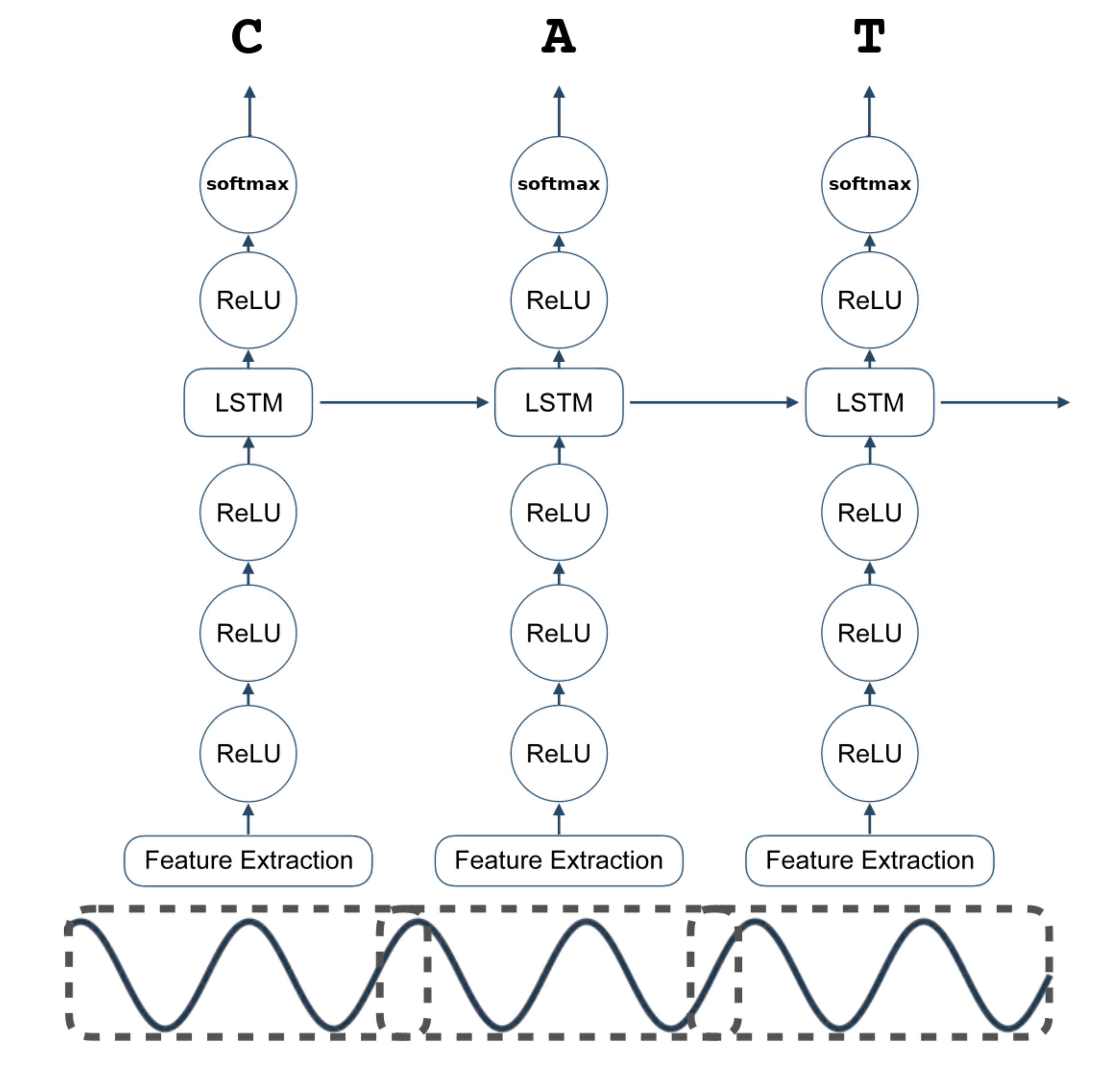}
	\caption{Architecture of Mozilla's DeepSpeech Automatic Speech Recognition model. A six-layer unidirectional CTC model, with one LSTM layer.}\label{fig:arch}
\end{figure}

All reported results were obtained with Mozilla's DeepSpeech v0.3.0 --- an open-source implementation of a variation of Baidu's first DeepSpeech paper~\cite{hannun14deep}.  This architecture is an end-to-end Automatic Speech Recognition (ASR) model trained via stochastic gradient descent with a Connectionist Temporal Classification (CTC) loss function \cite{graves2006connectionist}. The model is six layers deep: three fully connected layers followed by a unidirectional LSTM layer followed by two more fully connected layers (c.f. Figure (\ref{fig:arch})). All hidden layers have a dimensionality of 2,048 and a clipped ReLU activation. The output layer has as many dimensions as characters in the alphabet of the target language (including any desired punctuation as well as the blank symbol used for CTC). The input layer accepts a vector of 19 spliced frames (9 past frames + 1 present frame + 9 future frames) with 26 MFCC features each (i.e. a single, 494-dimensional vector).

All models were trained with the following hyperparameters on a single GPU. We use a batch-size of 24 for train 
and 48 for development, a dropout rate of 20\%, and a learning rate of $0.0001$ with the ADAM 
optimizer.\footnote{For a complete list of ADAM hyperparameters: \url{https://github.com/mozilla/DeepSpeech/blob/v0.3.0/DeepSpeech.py\#L79}} 
The new, target-language layers were initialized via Xavier initialization \cite{glorot2010understanding}. After every epoch of backpropagation over the training set, the loss over the entire development set is calculated. This development loss is used to trigger early stopping. Early stopping is triggered when the loss on the held-out development set either (1) increases over a window of five sequential epochs, or (2) the most recent loss over the development set has not improved in a window of five epochs more than a mean loss threshold of $0.5$ \textit{and} the window of losses shows a standard deviation of less than $0.5$.

\section{Results}

\begin{table}[!htbp]
\centering
\scalebox{0.8}{
\begin{tabular}{lcrrrrrrrrrrr}
  \toprule
  &  \multicolumn{6}{c}{\textbf{Character Error Rate}}\\
\multirow{3}{*}{\textbf{Lang.}} & \multicolumn{6}{c}{Number of Layers Copied from English} \\
                 & None & 1 & 2 & 3 & 4 & 5\\
\midrule
\texttt{sl}  & \cellcolor{gray!20.89720468890892}23.35 & \cellcolor{gray!32.394048692515796}21.65 & \cellcolor{gray!0.0}26.44 & \cellcolor{gray!49.70694319206493}19.09 & \cellcolor{gray!75.0}\textbf{15.35} & \cellcolor{gray!57.34896302975653}17.96 \\
\texttt{ga}  & \cellcolor{gray!3.843490304709192}31.83 & \cellcolor{gray!12.36149584487535}31.01 & \cellcolor{gray!0.0}32.2 & \cellcolor{gray!48.82271468144046}27.5 & \cellcolor{gray!70.42936288088642}25.42 & \cellcolor{gray!75.0}\textbf{24.98} \\
\texttt{cv}  & \cellcolor{gray!0.0}48.1 & \cellcolor{gray!3.5902345619913802}47.1 & \cellcolor{gray!12.637625658209679}44.58 & \cellcolor{gray!19.20775490665391}42.75 & \cellcolor{gray!75.0}\textbf{27.21} & \cellcolor{gray!58.01819052178075}31.94 \\
\texttt{br}  & \cellcolor{gray!0.0}21.47 & \cellcolor{gray!31.614963503649626}19.16 & \cellcolor{gray!19.981751824817486}20.01 & \cellcolor{gray!46.669708029197096}18.06 & \cellcolor{gray!75.0}\textbf{15.99} & \cellcolor{gray!41.74270072992698}18.42 \\
\texttt{tr}  & \cellcolor{gray!1.751373626373649}34.66 & \cellcolor{gray!7.314560439560452}34.12 & \cellcolor{gray!0.0}34.83 & \cellcolor{gray!31.31868131868132}31.79 & \cellcolor{gray!75.0}\textbf{27.55} & \cellcolor{gray!52.43818681318683}29.74 \\
\texttt{it}  & \cellcolor{gray!15.587595212187189}40.91 & \cellcolor{gray!1.3873775843308067}42.65 & \cellcolor{gray!0.0}42.82 & \cellcolor{gray!48.3949945593036}36.89 & \cellcolor{gray!75.0}\textbf{33.63} & \cellcolor{gray!63.00326441784549}35.10 \\
\texttt{cy}  & \cellcolor{gray!0.0}34.15 & \cellcolor{gray!31.1111111111111}31.91 & \cellcolor{gray!7.222222222222172}33.63 & \cellcolor{gray!55.83333333333334}30.13 & \cellcolor{gray!75.0}\textbf{28.75} & \cellcolor{gray!52.361111111111114}30.38 \\
\texttt{tt}  & \cellcolor{gray!0.0}32.61 & \cellcolor{gray!14.297253634894993}31.43 & \cellcolor{gray!21.930533117932143}30.80 & \cellcolor{gray!58.400646203554146}27.79 & \cellcolor{gray!75.0}\textbf{26.42} & \cellcolor{gray!48.22294022617127}28.63 \\
\texttt{ca}  & \cellcolor{gray!14.595375722543409}38.01 & \cellcolor{gray!55.05780346820808}35.21 & \cellcolor{gray!0.0}39.02 & \cellcolor{gray!54.33526011560696}35.26 & \cellcolor{gray!75.0}\textbf{33.83} & \cellcolor{gray!37.71676300578041}36.41 \\
\texttt{fr}  & \cellcolor{gray!42.18749999999989}43.33 & \cellcolor{gray!58.59374999999994}43.26 & \cellcolor{gray!0.0}43.51 & \cellcolor{gray!63.28124999999901}43.24 & \cellcolor{gray!72.6562499999988}43.20 & \cellcolor{gray!75.0}\textbf{43.19} \\
\texttt{kab}  & \cellcolor{gray!42.01570680628267}25.76 & \cellcolor{gray!52.225130890052384}25.5 & \cellcolor{gray!0.0}26.83 & \cellcolor{gray!62.04188481675397}25.25 & \cellcolor{gray!75.0}\textbf{24.92} & \cellcolor{gray!60.86387434554973}25.28 \\
\texttt{de}  & \cellcolor{gray!0.0}43.76 & \cellcolor{gray!32.81250000000083}43.69 & \cellcolor{gray!65.62500000000166}43.62 & \cellcolor{gray!75.0}\textbf{43.60} & \cellcolor{gray!0.0}43.76 & \cellcolor{gray!32.81250000000083}43.69 \\
\bottomrule
\end{tabular}
}
\caption{Fine-Tuned Transfer Learning Character Error Rate for each language, in addition to a baseline trained from scratch on the target language data. Bolded values display best model per language. Shading indicates relative performance per language, with darker indicating better models.}\label{table:results:tuned}
\end{table}

\begin{figure}[!htbp]
\centering
        \centering
\includegraphics[width=.48\textwidth]{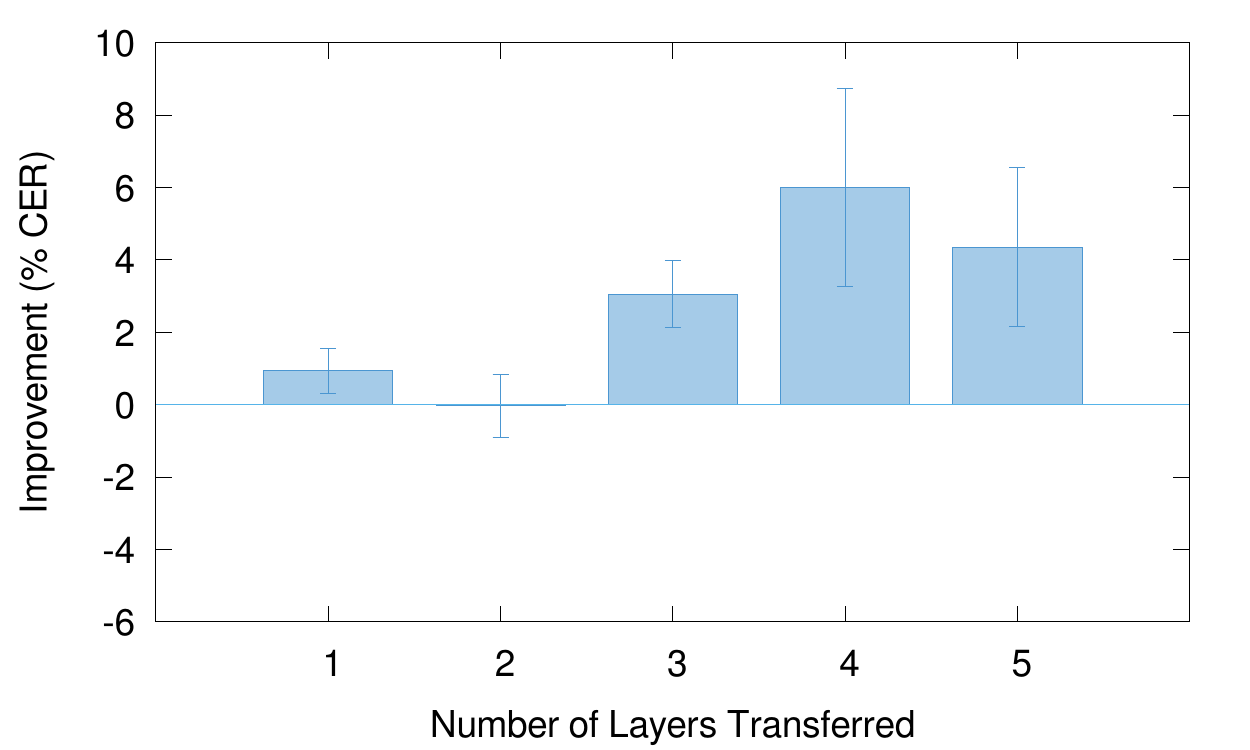}
\label{fig:layers:tuned}
	\caption{Mean and standard deviation in Character Error Rate improvement over all twelve languages investigated, relative to number of layers transferred from a pre-trained English DeepSpeech v0.3.0 model.}\label{fig:layers}
\end{figure}

The results from all experiments can be found in Table (\ref{table:results:tuned}). Each cell in the table contains the Character Error Rate (CER)\footnote{We report Character Error Rate as opposed to Word Error Rate (WER) because the former is more language-agnostic. Word Error Rate is most appropriate for languages which exhibit an analytic or isolating morphology and clearly delimit words in their orthography. Many languages do not use whitespace to delimit words, and there is no clear definition of ``word'' in a multilingual context. In short, both CER and WER make sense for languages like English, but CER is much more appropriate for languages like Mandarin, Turkish, and Chukchi.} of the resulting model on the test set, defined as the Levenshtein distance \cite{fiscus2006multiple} of the characters between the ground-truth transcript and the decoding result. The results in Table (\ref{table:results:tuned}) show how the number of layers transfered (columns) influence the performance on individual target languages (rows). Shaded cells indicate relative performance per language, where a darker cell represents a more accurate model. From this table we observe a trend in which four layers copied from pre-trained English DeepSpeech result in the best final model. This trend becomes more obvious in Figure (\ref{fig:layers}), where we average the improvement over all languages relative to the number of layers transfered from a source model.

\section{Concluding remarks}

We have presented Common Voice: a crowd-sourced, multilingual speech corpus which can scale to any language via community effort. All of the speech data is released under a Creative Commons CC0 license, making Common Voice the largest public domain corpus designed for Automatic Speech Recognition. In Section (\ref{sec:process}) we described the recording and validation process used to create the corpus. In Section (\ref{sec:contents}) we presented the current contents of Common Voice, and lastly in Section (\ref{sec:experiments}) we show multilingual Automatic Speech Recognition experiments using the corpus. There are currently 38 language communities collecting data via Common Voice, and we welcome more languages and more volunteers.

\section*{Acknowledgments}

Common Voice is a living project, and would not be possible without the thousands of hours given by volunteers. We thank all volunteers for their time, and especially the minority language activists who translate, find new texts, and organize Common Voice donation events. We thank George Roter, Gheorghe Railean, Rubén Martín, and Jane Scowcroft for their work on Common Voice, and all members of the Common Voice team, past and present.

This material is based upon work when Josh Meyer was supported by the National Science Foundation under Grant No. (DGE-1746060). Opinions, findings, conclusions, and recommendations are those of the authors and do not necessarily reflect the views of the NSF.

\section{References}
\label{main:ref}

\bibliographystyle{lrec}
\bibliography{paper}

\begin{thebibliography}{}

\bibitem[\protect\citename{Fiscus \bgroup et al.\egroup
  }2006]{fiscus2006multiple}
Fiscus, J.~G., Ajot, J., Radde, N., and Laprun, C.
\newblock (2006).
\newblock Multiple dimension levenshtein edit distance calculations for
  evaluating automatic speech recognition systems during simultaneous speech.
\newblock In {\em LREC}, pages 803--808. Citeseer.

\bibitem[\protect\citename{Gales \bgroup et al.\egroup }2014]{gales2014speech}
Gales, M.~J., Knill, K.~M., Ragni, A., and Rath, S.~P.
\newblock (2014).
\newblock Speech recognition and keyword spotting for low-resource languages:
  Babel project research at cued.
\newblock In {\em Spoken Language Technologies for Under-Resourced Languages}.

\bibitem[\protect\citename{Glorot and Bengio}2010]{glorot2010understanding}
Glorot, X. and Bengio, Y.
\newblock (2010).
\newblock Understanding the difficulty of training deep feedforward neural
  networks.
\newblock In {\em Proceedings of the thirteenth international conference on
  artificial intelligence and statistics}, pages 249--256.

\bibitem[\protect\citename{Graves \bgroup et al.\egroup
  }2006]{graves2006connectionist}
Graves, A., Fern{\'a}ndez, S., Gomez, F., and Schmidhuber, J.
\newblock (2006).
\newblock Connectionist temporal classification: labelling unsegmented sequence
  data with recurrent neural networks.
\newblock In {\em Proceedings of the 23rd international conference on Machine
  learning}, pages 369--376. ACM.

\bibitem[\protect\citename{Hannun \bgroup et al.\egroup }2014]{hannun14deep}
Hannun, A.~Y., Case, C., Casper, J., Catanzaro, B., Diamos, G., Elsen, E.,
  Prenger, R., Satheesh, S., Sengupta, S., Coates, A., and Ng, A.~Y.
\newblock (2014).
\newblock {Deep Speech}: Scaling up end-to-end speech recognition.
\newblock {\em CoRR}, abs/1412.5567.

\bibitem[\protect\citename{M-AILABS}2019]{mailabs}
M-AILABS.
\newblock (2019).
\newblock The m-ailabs speech dataset.
\newblock \url{https://www.caito.de/2019/01/the-m-ailabs-speech-dataset/}.
\newblock accessed 11/25/2019.

\bibitem[\protect\citename{Meyer}2019]{meyer2019multi}
Meyer, J.
\newblock (2019).
\newblock {\em Multi-Task and Transfer Learning in Low-Resource Speech
  Recognition}.
\newblock {PhD} dissertation, The University of Arizona.

\bibitem[\protect\citename{Roter}2019]{roter_2019}
Roter, G.
\newblock (2019).
\newblock Sharing our common voices – mozilla releases the largest to-date
  public domain transcribed voice dataset, Feb.
\newblock
  https://blog.mozilla.org/blog/2019/02/28/sharing-our-common-voices-mozilla-releases-the-largest-to-date-public-domain-transcribed-voice-dataset/.

\bibitem[\protect\citename{VoxForge}2019]{voxforge}
VoxForge.
\newblock (2019).
\newblock Voxforge.
\newblock \url{http://www.voxforge.org/}.
\newblock accessed 11/25/2019.

\end{thebibliography}

\end{document}